\documentclass[runningheads]{llncs}

\usepackage{eccv}

\usepackage{eccvabbrv}

\usepackage{graphicx}
\usepackage{booktabs}

\usepackage[accsupp]{axessibility}  

\usepackage[pagebackref,breaklinks,colorlinks,citecolor=eccvblue]{hyperref}

\usepackage{orcidlink}
\usepackage{url}
\usepackage{booktabs}
\usepackage{multirow}
\usepackage{multicol}
\usepackage{kotex}

\usepackage{algorithm}
\usepackage{listings}
\usepackage{xspace}
\usepackage{graphicx}
\usepackage{wrapfig}
\usepackage{subcaption}
\usepackage{pifont}

\definecolor{lightgreen}{RGB}{220,255,180}
\definecolor{darkergreen}{RGB}{21, 152, 56}
\newcommand\greenp[1]{\textcolor{darkergreen}{(#1)}}
\definecolor{red2}{RGB}{252, 54, 65}
\newcommand\redp[1]{\textcolor{red2}{(#1)}}
\newcommand\greenpscript[1]{\scriptsize\greenp{#1}}
\newcommand\redpscript[1]{\scriptsize\redp{#1}}

\begin{document}

\title{Rotary Position Embedding \\for Vision Transformer} 

\titlerunning{RoPE for ViT}

\author{Byeongho Heo\orcidlink{0000-0003-2134-6345} \quad Song Park\orcidlink{0000-0002-4332-9733} \quad Dongyoon Han\orcidlink{0000-0002-9130-8195} \quad Sangdoo Yun\orcidlink{0000-0002-0417-8450}}

\authorrunning{B.~Heo et al.}

\institute{NAVER AI Lab}

\maketitle

\begin{abstract}
Rotary Position Embedding (RoPE) performs remarkably on language models, especially for length extrapolation of Transformers. However, the impacts of RoPE on computer vision domains have been underexplored, even though RoPE appears capable of enhancing Vision Transformer (ViT) performance in a way similar to the language domain. This study provides a comprehensive analysis of RoPE when applied to ViTs, utilizing practical implementations of RoPE for 2D vision data. The analysis reveals that RoPE demonstrates impressive extrapolation performance, i.e., maintaining precision while increasing image resolution at inference. It eventually leads to performance improvement for ImageNet-1k, COCO detection, and ADE-20k segmentation. We believe this study provides thorough guidelines to apply RoPE into ViT, promising improved backbone performance with minimal extra computational overhead. Our code and pre-trained models are available at {\small \url{https://github.com/naver-ai/rope-vit}}

\end{abstract}

\section{Introduction}
\label{sec:intro}

Transformers~\cite{vaswani2017attention} have become popular due to their strong performance across various tasks in language and computer vision domains~\cite{devlin2018bert,dosovitskiy2020vit}. 
The transformer treats input data as a sequence of tokens. 
The tokens equally interact with others through a self-attention mechanism~\cite{vaswani2017attention}.
Since the self-attention mechanism is independent of the token index or positions (\ie, permutation invariance), the transformer requires additional position information, usually injected by position embedding~\cite{vaswani2017attention, shaw2018rpe, devlin2018bert, raffel2020rpe}.
The position embeddings give the position information to input tokens with specific embedding designed for the transformer.
They uniquely differentiate tokens based on their locations rather than their contents.
Thus, the position information of self-attention heavily depends on the position embedding, which is a crucial component in designing transformer architectures.

There are two primary methods in position embedding for Vision Transformers: Absolute Positional Embedding (APE) \cite{devlin2018bert,dosovitskiy2020vit} and Relative Position Bias (RPB) \cite{shaw2018rpe,raffel2020rpe,liu2021swin}. APE utilizes the absolute position of tokens for position embedding through sinusoidal or learnable embedding. Otherwise, RPB enables relative positions between tokens by adding relative biases to the attention matrix of the self-attention layers. In general, APE is used for traditional ViT architecture~\cite{dosovitskiy2020vit}, and RPE is preferred to hierarchical ViT like Swin Transformer~\cite{liu2021swin}. Although both position embeddings are effective for the transformer on fixed-resolution settings, they struggle with resolution changes, requiring flexibility and extrapolation in position embeddings. Considering that the resolution of pre-training is usually smaller than that of downstream dense prediction, it might degrade ViT performance in various applications, such as multi-resolution recognition, object detection, and segmentation.

This paper aims to improve position embedding for vision transformers by applying an extended Rotary Position Embedding (RoPE) \cite{su2024roformer}.
RoPE is a relative position embedding that is specially designed for extrapolation in language domains.
Despite the remarkable success of RoPE in Large-Language Models~\cite{touvron2023llama2,roziere2023codellama,jiang2023mistral}, its effectiveness in vision tasks has not been validated due to limited investigation.
In this paper, we provide a comprehensive investigation of RoPE for transformers in vision recognition tasks.
Our investigation starts with 1D to 2D expansion of RoPE to cope with images rather than original language inputs.
Although 2D RoPE using axial frequencies was used in pioneer works~\cite{lu2023unified,fang2023eva,lu2024fit}, we argue that it lacks the ability to handle diagonal directions, which are preferred in convolution networks by the square kernel.
To cope with the diagonal direction of RoPE, we propose to use mixed axis frequencies for 2D RoPE, named RoPE-Mixed.
Since RoPE-Mixed uses frequencies for both axes as learnable network parameters, it effectively handles diagonal direction and is more suitable for ViT's attention than Axial 2D RoPE.

In experiments, we apply variants of 2D RoPE to representative transformer architectures, ViT and Swin Transformer, and validate the effects of 2D RoPE in various tasks, including multi-resolution classification on ImageNet-1k~\cite{deng2009imagenet}, object detection on MS-COCO~\cite{lin2014mscoco}, and semantic segmentation on ADE20k~\cite{zhou2017ade1,zhou2019ade2}. The results show that 2D RoPE is a beneficial option for position embedding in transformers with impressive performance improvements on high-resolution images, i.e., extrapolation of images. We believe our study demonstrates the significant impact of 2D RoPE in vision domains and contributes to future research by suggesting a beneficial option in position embedding for vision transformers. 
\section{Related Works}
\label{sec:relworks}

\subsubsection{Position embedding.}
ViT~\cite{dosovitskiy2020vit} introduces a transformer~\cite{vaswani2017attention} architecture for visual inputs, employing Absolute Positional Embedding (APE)~\cite{devlin2018bert,dosovitskiy2020vit}. APE with learnable parameters effectively injects spatial positions of each token to be used for the self-attention mechanism. Hierarchical ViT such as Swin Transformer~\cite{liu2021swin} increase the spatial length of tokens at early layers using pooling. To handle a large number of tokens with limited position embeddings, Relative Position Bias (RPB)~\cite{shaw2018rpe,raffel2020rpe,liu2021swin} is preferred by the hierarchical ViTs.
Studies have been conducted to improve position embedding for ViT based on these two major position embeddings.
iRPE~\cite{wu2021rethinking} proposes an improved RPB by applying relative position embedding as multiplication with query vector.
CPE~\cite{chu2021conditional} finds that a convolution network can effectively inject relative position information to tokens and utilizes $3 \times 3$ depth-wise convolution~\cite{mobilenetv1} as conditional position embedding.
LaPE~\cite{yu2023lape} shows that simple scaling with adaptive layer-norm can improve the positional embedding of various networks.

\vspace{-1em}
\subsubsection{RoPE in vision modeling.}
Pioneering studies introduced RoPE to ViT-related architectures. Hybrid X-former~\cite{jeevan2022resource} applies 1D RoPE to ViT variants named Vision X-formers; it is the first attempt at the application of RoPE in ViT to our knowledge.  However, 1D RoPE is insufficient to demonstrate performance, and evaluation is limited to small datasets such as CIFAR~\cite{cifar} and Tiny ImageNet~\cite{le2015tiny}.
EVA-02~\cite{fang2023eva} introduces 2D Axial RoPE to a new language-aligned vision model EVA-02, like  CLIP~\cite{clip}. 
Unified-IO 2~\cite{lu2023unified} uses 2D RoPE for new multi-modal modeling; 2D Axial RoPE is applied to non-text modalities, including vision, audio, and video. 
In diffusion modeling~\cite{rombach2022latent}, FiT~\cite{lu2024fit} applies 2D Axial RoPE for their new diffusion model.
In these studies, 2D Axial RoPE was used to improve new model performance on language-related or generation tasks, which differs from our goal of challenging classification, detection, and segmentation tasks. 
Exploring the impacts of 2D RoPE implementations in basic architectures with general training recipes could benefit diverse vision researchers.

\vspace{-1em}
\subsubsection{Multi-resolution inference.}
Unlike ConvNets~\cite{he2016resnet}, ViT~\cite{dosovitskiy2020vit} requires a transformation in position embedding for multi-resolution inference. 
Some studies investigated a multi-resolution inference method for ViT.
CAPE~\cite{likhomanenko2021cape} analyzes ViT's position embedding in resolution changes and finds that augmenting position embedding improves the multi-resolution performance of ViT.
Thus, they propose a new training recipe that includes continuous augmenting of position embedding (CAPE).
ResFormer~\cite{tian2023resformer} shows that relative position embedding based on depth-wise convolution layer benefits multi-resolution inference. Using this property, the study proposes an improved ViT architecture with global and local depth-wise conv embedding. It substantially improves multi-resolution performance with multi-resolution self-distillation learning recipes.
In contrast to conventional multi-resolution, FlexiViT~\cite{beyer2023flexivit} proposes a ViT with flexible patch sizes that can replace multi-resolution inference. In FlexiViT, ViT increases the patch size instead of increasing input resolution. By training with a multi-patch-size training scheme and distillation using ViT-B/8~\cite{steiner2021augreg}, FlexiViT exhibits remarkable performance for various patch-size, which corresponds to multi-resolution in computation cost aspect.

These studies require special training methods, which make them difficult to combine with other training recipes, potentially reducing general applicability.
RoPE improves multi-resolution performance while using existing training recipes as is, offering generally applicable and easy-to-use compared to others.
\section{Method}
\label{sec:method}
Rotary Position Embedding (RoPE)~\cite{su2024roformer} was introduced to apply to key and query in self-attention layers as channel-wise multiplications, which is distinct from conventional position embeddings - APE is added to the stem layer; RPB is added to an attention matrix. We first present conventional position embeddings, including RoPE~\cite{su2024roformer} in language model at \S\ref{sec:pos_embed}, and provide feasible expansion of RoPE to 2D inputs for transformers in the vision domain in subsequent \S\ref{subsec:RoPE2D}. In \S\ref{sec:discussion}, we describe the characteristics of RoPE compared to other position embedding and analysis for 2D RoPE.

\subsection{Preliminary: Introducing Position Embeddings}
\label{sec:pos_embed}

\subsubsection{Absolute Positional Embedding (APE)\cite{vaswani2017attention,devlin2018bert,dosovitskiy2020vit}}
is the most common position embedding for Vision Transformer (ViT). APE is generally added to the feature right after the patchification layer computes tokens from $16 \times 16$ or $32 \times 32$ patch images. For patch tokens $\mathbf{x}_0 \in \mathbb{R}^{N\times d}$, APE $\mathbf{E}_{APE} \in \mathbb{R}^{N\times d}$ gives the position information for each token by addition: 
\begin{equation}
\mathbf{x}'_0 = \mathbf{x}_0 + \mathbf{E}_{APE}.
\label{eq:ape_add}
\end{equation}
The tokens with APE $\mathbf{x}'_0$ are fed to transformer blocks and utilized as a feature merged with the absolute positional information. There are two variants on how to build $\mathbf{E}_{APE}$: sinusoidal and learnable embedding. Sinusoidal embedding uses axial sinusoidal functions as APE. When APE for position $\mathbf{p}_n = (p^x_n, p^y_n)$ is denoted as $\mathbf{E}_{APE}(\mathbf{p}_n) \in \mathbb{R}^d$, $t$-th dim of sinusoidal embedding $\mathbf{E}_{APE}(\mathbf{p}_n, t)$ is
\vspace{-1em}
\begin{align}
&\mathbf{E}_{APE}(\mathbf{p}_n, 4t) = \mathrm{sin}(p_n^x / 10^{4t/\lfloor \frac{d}{4} \rfloor}), \,
\mathbf{E}_{APE}(\mathbf{p}_n, 4t+1) =\mathrm{cos}(p_n^x / 10^{4t/\lfloor \frac{d}{4} \rfloor}),
\label{eq:sin_APE}
\\
&\mathbf{E}_{APE}(\mathbf{p}_n, 4t+2) = \mathrm{sin}(p_n^y / 10^{4t/\lfloor \frac{d}{4} \rfloor}), \,
\mathbf{E}_{APE}(\mathbf{p}_n, 4t+3) = \mathrm{cos}(p_n^y / 10^{4t/\lfloor \frac{d}{4} \rfloor}). \notag
\end{align}
Note that we use 0-base numbers for indexes $p^x_n, p^y_n$, and $t$.
The other implementation of APE is to use learnable parameters and train them with the training process.
$N \times d$ learnable parameters are randomly initialized and are used as Eq.~\ref{eq:ape_add}.
It is the simplest way for APE, and supervised learning recipes use APE with learnable parameters~\cite{dosovitskiy2020vit,touvron2021deit,touvron2022deit3}.
Since learnable APE is commonly used for ViT, we refer to it as the default option for APE.

\vspace{-1em}
\subsubsection{Relative Position Bias (RPB)}\cite{raffel2020rpe,liu2021swin} is a popular way to inject relative distances to the ViT architectures.
APE is not suitable for handling tokens based on their relative positions, as it relies solely on absolute positions in the image $\mathbf{p}_n = (p^x_n, p^y_n)$. 
It is necessary to use different types of position embedding that utilize relative positions $\tilde{\mathbf{p}}_{nm} = (\tilde{p}^x_{nm}, \tilde{p}^y_{nm}) = (p^x_n - p^x_m, p^y_n - p^y_m)$.
RPB is widely used relative position embedding for ViT. In contrast to learnable APE, which has learnable parameters for each absolute position, RPB uses learnable parameters for each relative position. i.e., Relative Position Bias (RPB) table $T$ is defined as learnable parameters for every possible relative position:
\begin{equation}
    T = \left\{ T_{\tilde{p}^x \tilde{p}^y} \in \mathbb{R} \mid \tilde{p}^x \in \{-W, \ldots, 0, \ldots, W\}, \tilde{p}^y \in \{-H, \ldots, 0, \ldots, H\} \right\}.
\label{eq:rpb_table}
\end{equation}
While APE is added to network features, RPB is directly applied to the attention matrix of every self-attention layer since it is the only position that can handle relative relations in transformer architecture. The attention matrix $\mathbf{A}\in \mathbb{R}^{N \times N}$ with the query and key of a head denoted by $\mathbf{q}, \mathbf{k} \in \mathbb{R}^{N \times d_{head}}$, is calculated 
\begin{equation}
    \mathbf{A} = \mathrm{SoftMax}(\mathbf{q} \mathbf{k}^T / \sqrt{d_{head}}).
    \label{eq:attention}
\end{equation}
To fit with the attention matrix, the RPB table $T \in \mathbb{R}^{2W \times 2H}$ is rearranged to RPB embedding $\mathbf{E}_{RPB} \in \mathbb{R}^{N \times N}$ where $(n, m)$-th component $\mathbf{E}^{RPB}_{nm}$ is
\begin{equation}
    \mathbf{E}^{RPB}_{nm} = T_{\tilde{p}^x_{nm} \tilde{p}^y_{nm}} = T_{(p^x_{n} - p^x_{m}) (p^y_{n} - p^y_{m})}.
\end{equation}
Then, RPB is added to the attention matrix in Eq.~\ref{eq:attention} as
\begin{equation}
    \mathbf{A} = \mathrm{SoftMax}(\mathbf{q} \mathbf{k}^T / \sqrt{d_{head}}) + \mathbf{E}_{RPB}.
\end{equation}
By RPB, self-attention handles relative positions. Note that we describe RPB for a head in a multi-head self-attention layer. Thus, in practice, RPB parameters and addition are repeated for each head in multi-head attention.

\vspace{-1em}
\subsubsection{Rotary Position Embedding (RoPE)}\cite{su2024roformer} is a recent method in the line of relative position embedding studies. Although RPB delivers relative position to the attention, simple addition as bias may limit interaction with attention weights, which causes limited utilization of relative position. Thus, RoFormer~\cite{su2024roformer} proposes a novel relative position embedding method: Rotary Position Embedding (RoPE). Note that this section explains the original RoPE designed for language modeling. We will explain our RoPE for 2D images in \S\ref{subsec:RoPE2D}

Limitations of RPB emerge from the addition to the attention matrix.
Since RPB is applied to the attention matrix after query-key multiplication, it cannot affect and contribute to the query-key similarity, which is the core operation of self-attention.
To resolve this limitation, RoPE introduces the multiplication of Euler's formula ($e^{i \theta}$) to key and query vectors as relative position embedding. i.e., when $n, m$-th query and key is $\mathbf{q}_n, \mathbf{k}_m \in \mathbb{R}^{1 \times d_{head}}$, RoPE is applied as
\begin{equation}
\mathbf{q}_n' = \mathbf{q}_n e^{i n \theta}, \; \mathbf{k}_m' = \mathbf{k}_m e^{i m \theta}.
\label{eq:rope_qk}
\end{equation}
Then, $(n, m)$-th component of attention matrix is calculated as
\begin{equation}
\mathbf{A}_{(n, m)}' = \mathrm{Re}[\mathbf{q}_n' \mathbf{k}_m'^*] = \mathrm{Re}[\mathbf{q}_n \mathbf{k}_m^* e^{i (n - m) \theta}],
\label{eq:rope_attn}
\end{equation}
where $\mathrm{Re}[\cdot]$ denotes real part of complex number and $^*$ means complex conjugates.
By multiplying complex rotation $e^{i \theta n}, e^{i \theta m}$ depending on token position $(n, m)$, RoPE injects relative positions $(n-m)$ to the attention matrix in rotation form. 
In practical implementation, RoPE converts $\mathbf{q}_n, \mathbf{k}_m \in \mathbb{R}^{1 \times d_{head}}$ to complex vector $\mathbf{\bar{q}}_n, \mathbf{\bar{k}}_m \in \mathbb{C}^{1 \times (d_{head}/2)}$ by considers ($2t$)-th dim as real part and ($2t+1$)-th dim as imaginary part. It produces the same attention value as $\mathbf{q}_n \mathbf{k}_m^T = \mathrm{Re}[\mathbf{\bar{q}}_n \mathbf{\bar{k}}_m^*]$ but reduces computational wastes. Also, RoPE utilizes multiple frequencies $\theta_t$ using channel dimensions of key and query as
\begin{equation}
    \theta_t = 10000^{-t / (d_{head}/2)},\;\mathrm{where}\, t \in \{0, 1, ... ,d_{head}/2\}.
\label{eq:rope_freqs}
\end{equation}
In summary, a rotation matrix $\mathbf{R} \in \mathbb{C}^{N \times (d_{head}/2)}$ is defined as
\begin{equation}
    \mathbf{R}(n, t) = e^{i \theta_t n}
\label{eq:rotation_matrix}
\end{equation}
and applied to query and key with the Hadamard product $\circ$ as
\begin{equation}
    \mathbf{\bar{q}}' = \mathbf{\bar{q}} \circ \mathbf{R}, \;\;\; \mathbf{\bar{k}}' = \mathbf{\bar{k}} \circ \mathbf{R}, \;\;\; \mathbf{A}' = \mathrm{Re}[\mathbf{\bar{q}}' \mathbf{\bar{k}}'^*].
\label{eq:rope_rotation_qk}
\end{equation}
Note that the attention matrix with RoPE $\mathbf{A}'$ implies relative position in rotation form $e^{i (n - m) \theta_t}$ for ($d_{head}/2$) number of frequencies, which gives a lot of performance beneficial to the transformer, especially for extrapolation on inference stage based on periodic functions.

\subsection{RoPE for 2D images}
\label{subsec:RoPE2D}

RoPE exhibits remarkable performance in the language domain. 
However, only a few studies have explored using RoPE in the vision domain with 2D input, as it was designed solely for 1D input.
This section introduces feasible implementations of 2D RoPE for input images: axial and learnable frequency.

\vspace{-1em}
\subsubsection{Axial frequency.} A typical way to expand 1D position embedding to 2D is repeating 1D operation for each axis. Similar to 2D sinusoidal embedding in Eq.~\ref{eq:sin_APE}, axial frequency is to divide embedding dimensions into two and apply position embedding for the x-axis and y-axis separately. It is straightforward because it is technically the same as repeating 1D embedding twice.

First, we need to change the 1D token index $n$ in RoPE to a 2D token position $\mathbf{p}_n = (p^x_n, p^y_n)$ where $p^x_n \in \{0,1, ... ,W\}$, $p^y_n \in \{0,1, ... ,H\}$ for token width $W$ and height $H$. Thus, the rotation matrix $\mathbf{R} \in \mathbb{C}^{N \times (d_{head}/2)}$ in Eq.~\ref{eq:rotation_matrix} is changed as
\begin{equation}
\mathbf{R}(n, 2t) = e^{i\theta_t p^x_n}, \;\; \mathbf{R}(n, 2t+1) = e^{i\theta_t p^y_n}.
\label{eq:rotation_axial}
\end{equation}
Also, the range of position indexes $(p^x_n, p^y_n)$ is reduced by square root. It is natural to reduce RoPE frequencies $\theta_t$ in Eq.~\ref{eq:rope_freqs} by square root as
\begin{equation}
    \theta_t = 100^{-t / (d_{head}/4)},\;\mathrm{where}\, t \in \{0, 1, ... ,d_{head}/4\}.
\end{equation}
Note that $\theta_t$ for vision is often larger than that of language, and the number of frequencies is halved to cover both (x, y) dimensions with $d_{head}$ as well. This axial frequency has been used in a few pioneering works~\cite{fang2023eva,lu2023unified,lu2024fit} to further improve the performance of a new ViT architecture.

\vspace{-1em}
\subsubsection{Mixed learnable frequency.}
The axial frequency is a simple but effective way to expand RoPE for the vision domain.
However, it is unable to handle diagonal directions since the frequencies only depend on a single axis.
RoPE injects relative positions in the form of Euler's formula ($e^{i\theta_t (n - m)}$). 
Thus, with axial frequencies, the relative positions are applied as axial directions $e^{i\theta_t (p^x_n - p^x_m)}$ or $e^{i\theta_t (p^y_n - p^y_m)}$, which cannot be converted to mixed frequency $e^{i (\theta^x_t \tilde{p}^x_{nm} + \theta^y_t \tilde{p}^y_{nm})}$.
In the case of sinusoidal APE in Eq.~\ref{eq:sin_APE}, the sinusoidal functions can be mixed with another axis through query-key multiplication in the self-attention layer. 
However, RoPE already spends query-key multiplication for position subtraction for relative distance.
There is no way to mix axial frequencies for diagonal direction.
We conjecture that it might degrade RoPE's potential performance and make sub-optimal axial frequency choices in the vision domain.

To handle mixed frequencies, we propose to use a rotation matrix in Eq.~\ref{eq:rotation_matrix} in mixed axis form as
\begin{equation}
    \mathbf{R}(n, t) = e^{i (\theta^x_t p^x_n + \theta^y_t p^y_n)}.
\label{eq:rotation_mixed}
\end{equation}
By using two frequencies for each axis, RoPE allows handling of the diagonal axis. 
The RoPE attention matrix in Eq.~\ref{eq:rope_attn} is changed by mixed frequency as
\begin{equation}
\mathbf{A}_{(n, m)}' = \mathrm{Re}[\mathbf{q}_n \mathbf{k}_m^* e^{i (\theta^x_t (p^x_n - p^x_m) + \theta^y_t (p^y_n - p^y_m))}].
\label{eq:rope_attn_mixed}
\end{equation}
This formulation is identical to the axial frequency implementation as $\theta^x_t$ or $\theta^y_t$ goes to zero. Thus, mixed frequency RoPE is a generalized version of axial frequency RoPE. 
Different from fixed frequencies in language RoPE and axial frequency, we let the network learn frequencies $(\theta^x_t, \theta^y_t)$ for $t \in \{0, 1, ...,d_{head}/2\}$ as learnable parameters.
Our mixed learnable frequency implementation enables diagonal direction handling to RoPE and makes RoPE learnable, like conventional positional embedding in the vision domain. Like RPB, we use separate sets of learnable frequencies for each head and every self-attention layer.
It produces $d$ learnable parameters per self-attention layer. However, it is negligible since it requires only $\sim$0.01\% of network parameters in ViT-B.

\subsection{Discussion}
\label{sec:discussion}

\subsubsection{2D Fourier analysis.} 
\begin{figure}[t]
    \centering
    \includegraphics[width=1.0\linewidth]{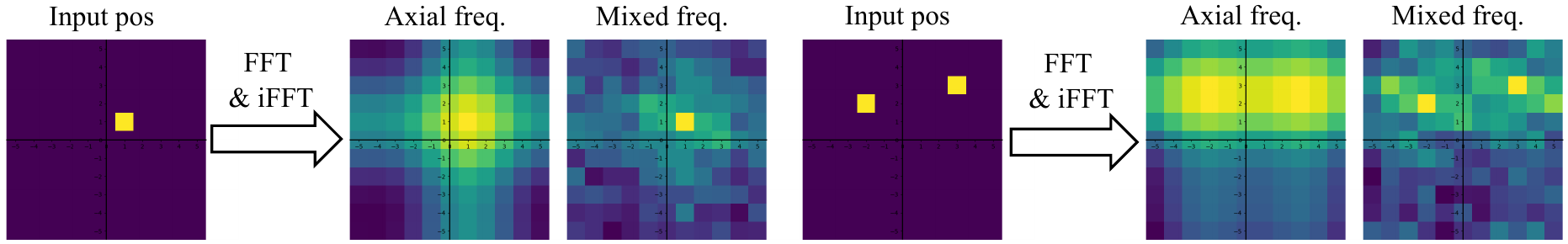}
    \vspace{-1em}
    \caption{\small \textbf{2D Fourier reconstruction with RoPE frequencies.} We perform a Fast Fourier Transform (FFT) followed by an inverse FFT with only RoPE frequencies to evaluate the representation capabilities of RoPE frequencies}
    \label{fig:2d_fourier}
    \vspace{-0.4cm}
\end{figure}

We design a 2D Fourier analysis to demonstrate the representational difference between RoPE-Axial and RoPE-Mixed. When all 2D frequencies are utilized, a 2D Fast Fourier Transform (FFT) followed by an inverse Fast Fourier Transform (iFFT) perfectly reconstructs the input. 
However, the number of RoPE frequencies is limited to $\frac{d_{head}}{2}$ as in Eq.~\ref{eq:rope_freqs}. 
$\frac{d_{head}}{2}$ ($= 32$ for ViT-B) frequencies are insufficient to cover all 2D frequencies, resulting in imperfect reconstructions. 
This imperfect reconstruction reflects the expressiveness and representation pattern of the frequencies.
In Fig.~\ref{fig:2d_fourier}, we compare 2D FFT-iFFT results of RoPE-Axial and RoPE-Mixed frequencies. Note that we use RoPE-Mixed frequencies from ViT-B trained on ImageNet-1k.
The results show a significant difference: Axial frequencies exhibit artifacts along axial lines, impairing precise positional representation, whereas Mixed frequencies utilize diverse 2D frequencies to produce sharper locations.
Thus, we claim that the mixed frequencies are necessary for precise localization in the attention, which explains why RoPE-Mixed performs better than RoPE-Axial in \S\ref{sec:experiment}.

\vspace{-1em}
\subsubsection{On image resolution changes.} Vision models use diverse image resolutions depending on the goal of target tasks. For example, image classification uses $224 \times 224$ as the standard resolution for comparison but utilizes small resolutions~\cite{touvron2022deit3,wightman2021resnet} for training efficiency and enlarges resolutions to boost the performance additionally. Furthermore, object detection and segmentation prefer larger resolutions to capture small objects. Thus, transformers for vision should support resolution changes, which is linked to the necessity of resolution change in position embedding.
RoPE makes an extended position embedding based on sinusoidal function for large resolution. Different from zero-padding in RPB, the rotation matrix in Eq.~\ref{eq:rotation_axial} and Eq.~\ref{eq:rotation_mixed} can produce values for extended positions since it is based on periodic functions, which has proven its effectiveness for extrapolation~\cite{touvron2023llama2,roziere2023codellama}.
We expect that the advantage of RoPE in extrapolation will also be effective for multi-resolution benchmark in \S\ref{subsec:exp_multi_res} and dense prediction tasks in \S\ref{subsec:exp_det} and \S\ref{subsec:exp_seg}.

\vspace{-1em}
\subsubsection{Phase shift in RoPE.}
In sinusoidal representation, phase shift such as $\phi$ in $e^{i (m-n)\theta + i\phi}$ is an important ability to control activation area.
This phase shift ability is already included in $\mathbf{W}_q$ and $\mathbf{W}_k$ of the self-attention layer.
Based on Eq.~\ref{eq:rope_attn}, when we apply $e^{i (m-n)\theta + i\phi}$ and $\mathbf{q}_n = \mathbf{x}_n \mathbf{W}_q $, the equation is
\begin{equation}
    \mathbf{x}_n \mathbf{W}_q e^{i (n - m) \theta + i\phi}\mathbf{k}_m^* = \mathbf{x}_n \mathbf{W}_q e^{i\phi}\mathbf{k}_m^* e^{i (n - m) \theta} = \mathbf{x}_n \mathbf{W}_q' \mathbf{k}_m^* e^{i (n - m) \theta}.
\end{equation}
Thus, RoPE does not need additional parameters for phase shift $\phi$ since learnable parameters $\mathbf{W}_q$ and $\mathbf{W}_k$ can do the same role in network training.


\vspace{-1em}
\subsubsection{Analyzing attention.}
\begin{figure}[t]
    \centering
    \includegraphics[width=0.95\linewidth]{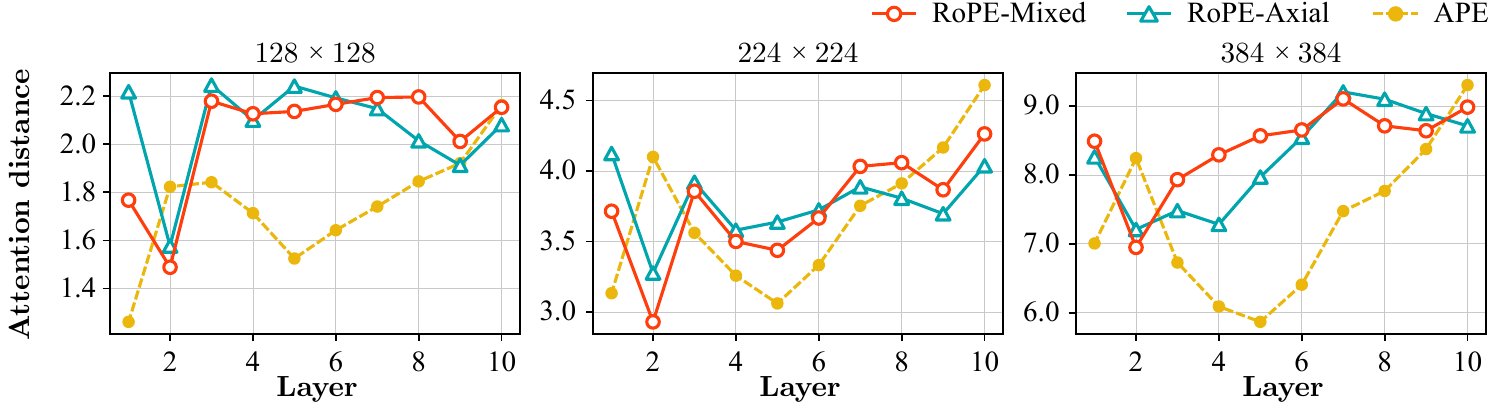}
    \vspace{-1em}
    \caption{\small \textbf{Attention distances of ViT-B for APE/RoPE.} We measure the average distance of attention interaction by computing the distance between query-key tokens from attention probabilities. We average the distance across the validation set. }
    \label{fig:attn_distance}
    \vspace{-0.2cm}
\end{figure}
\begin{figure}[t]
    \centering
    \includegraphics[width=0.95\linewidth]{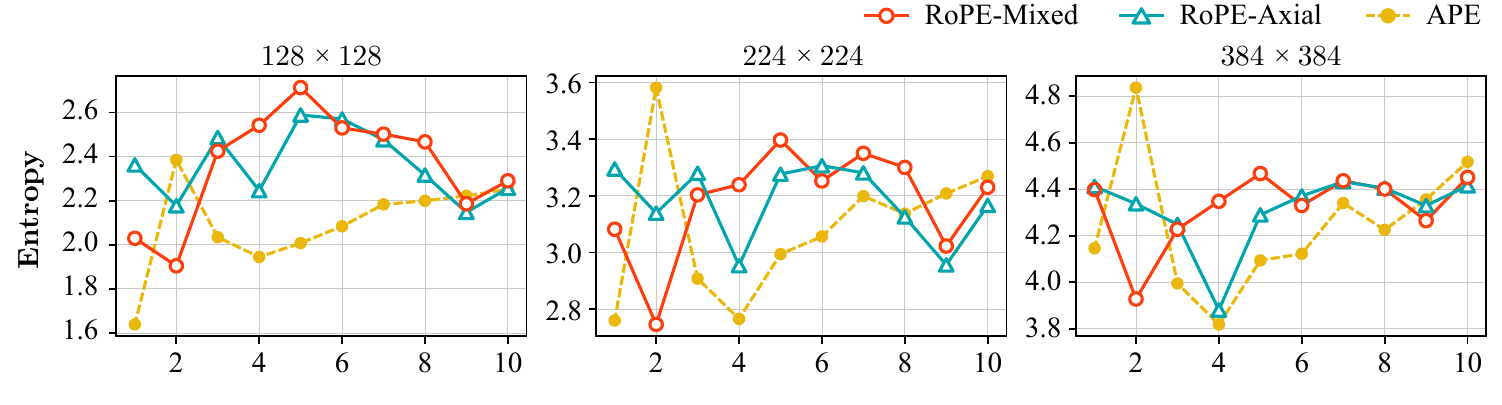}
    \vspace{-1em}
    \caption{\small \textbf{Entropy of attention in ViT-B with APE or RoPE.} Entropy of attention probability is measured for every self-attention of ViT-B. A high entropy value indicates that a large number of tokens are involved in the attention interaction.}
    \label{fig:attn_entropy}
    \vspace{-0.4cm}
\end{figure}

We analyze the attention matrix of RoPE ViT compared to the ViT with APE. Following attention analysis in literature~\cite{heo2021rethinking,park2023self}, we measure attention distances and entropy on the ImageNet-1k validation set with various resolutions. 
Attention distance refers to the average spatial distance involved in attention interaction. Attention entropy represents the entropy values of attention probabilities, indicating the sharpness of attention.
The averaged attention distances are shown in Fig.~\ref{fig:attn_distance}. In training resolution $224 \times 224$, RoPEs increase attention distance at the middle layers but decrease it in the second and later layers. In other resolutions, the pattern is similar, but the difference is more significant than the training resolution. In short, RoPEs increase attention distance at the middle layers, which becomes substantial at resolution changes. The entropy results are reported in Fig.~\ref{fig:attn_entropy}. Interestingly, the pattern is similar to attention distance. Entropy of RoPE is larger than that of APE at the middle layers. These analysis results imply that RoPE makes attention interact with long-range (attention distance) and various tokens (entropy). We speculate that these differences in attention contributed to the performance improvement of RoPE observed in \S\ref{sec:experiment}.


\vspace{-1em}
\subsubsection{Computation costs.} Although RoPE has an involved formulation compared with APE and RPB, its computation cost is negligible to the overall computation. The rotation matrix in Eq.~\ref{eq:rotation_axial} and \ref{eq:rotation_mixed} is pre-computed before inference. The Hadamard product in Eq.~\ref{eq:rope_rotation_qk} is the only computation required for inference - 1.8M FLOPs for ViT-B and accounts for only 0.01\% of ViT-B's 17.6G FLOPs.
\section{Experiments}
\label{sec:experiment}

We apply 2D RoPE to two representative ViT architectures: ViT~\cite{dosovitskiy2020vit} and Swin Transformer~\cite{liu2021swin}.
Note that ViT uses APE, whereas Swin Transformer uses RPB. Thus, our experiment can verify the performance of RoPE when it replaces APE or RPB.
RoPE in ViT and Swin Transformer is validated for image recognition, including multi-resolution classification (\S\ref{subsec:exp_multi_res}) on ImageNet-1k~\cite{deng2009imagenet}, object detection (\S\ref{subsec:exp_det}) on MS-COCO~\cite{lin2014mscoco}, and semantic segmentation (\S\ref{subsec:exp_seg}) on ADE20k~\cite{zhou2017ade1,zhou2019ade2}. We compare the conventional position embeddings (APE, RPB) with two variants of 2D RoPE RoPE-Axial (Eq.~\ref{eq:rotation_axial}) and RoPE-Mixed (Eq.~\ref{eq:rotation_mixed}). Our experiments will exhibit the remarkable performance of 2D RoPE across all tasks, particularly with a significant margin in extrapolation.

\subsection{Multi-resolution classification}
\label{subsec:exp_multi_res}

Robustness on multi-resolution inputs is an essential factor of ViT performance, as it is closely related to their downstream performance in dense prediction tasks. In language models~\cite{touvron2023llama2,roziere2023codellama,jiang2023mistral}, RoPE exhibited strong extrapolation performance, i.e., text sequence longer than training samples. 2D RoPE might be suitable for large-resolution images, leveraging its extrapolation capabilities. We train ViTs and Swin Transformers on ImageNet-1k~\cite{deng2009imagenet} training set with high-performance training recipes~\cite{touvron2022deit3,liu2021swin}. We report the accuracy on the ImageNet-1k validation set as varying image sizes. Note that we use the ImageNet-1k standard image resolution $224 \times 224$ for training. Thus, a resolution larger than 224 can be considered as extrapolation.

\vspace{-1em}
\subsubsection{Vision Transformer (ViT).} 

\begin{figure}[t]
    \centering
    \includegraphics[width=0.95\linewidth]{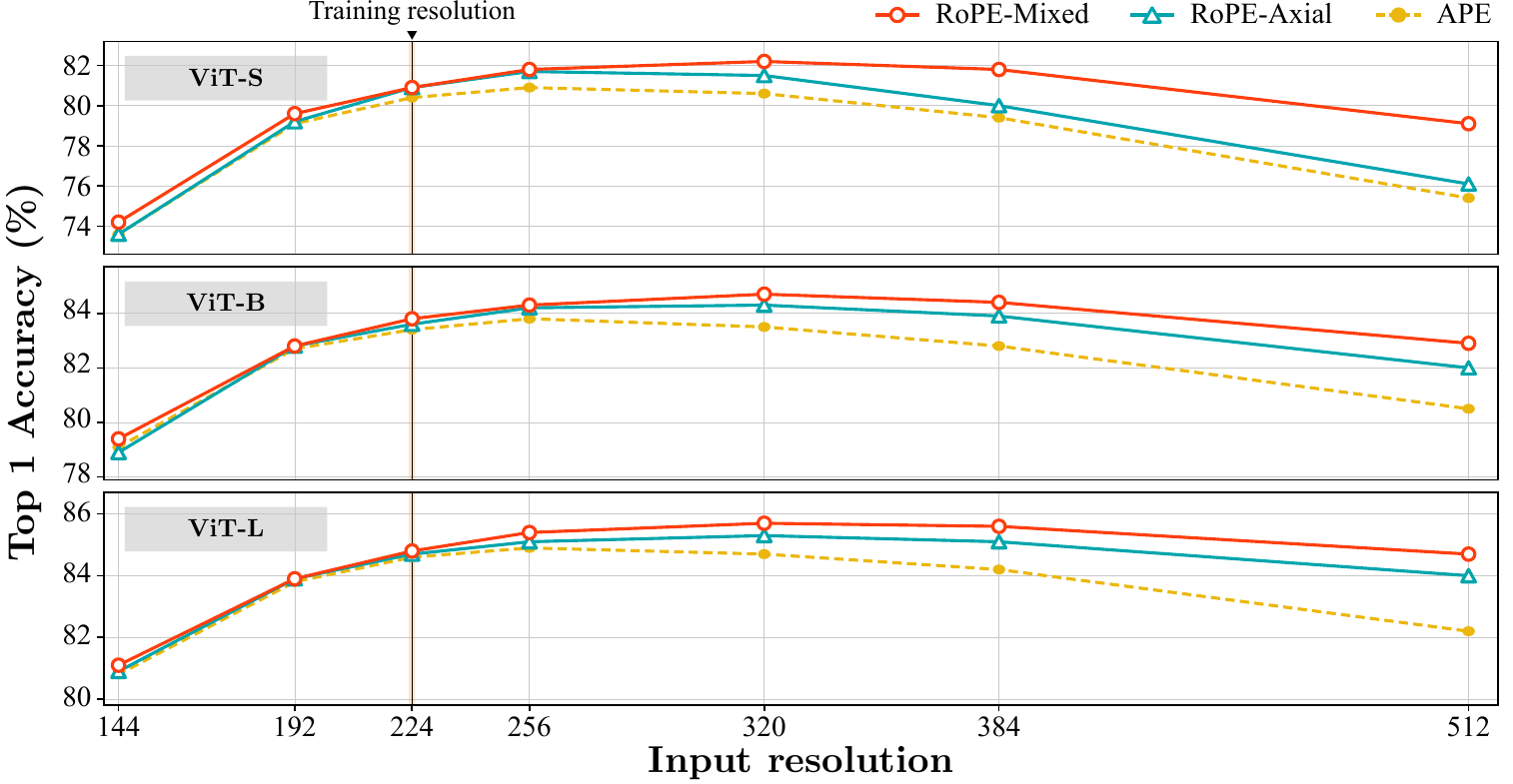}
    \vspace{-1em}
    \caption{\small \textbf{Multi-resolution performance of ViTs.} We apply two variants of 2D RoPE, RoPE-Axial, and RoPE-Mixed, to the ViT architectures. All ViTs are trained on ImageNet-1k~\cite{deng2009imagenet} with DeiT-III~\cite{touvron2022deit3}'s 400 epochs training recipe.}
    \label{fig:exp_multi_vit}
    \vspace{-0.4cm}
\end{figure}

We apply 2D RoPE to ViT-S, ViT-B, and ViT-L.
We train ViT with a strong supervised learning training recipe for ImageNet-1k, DeiT-III 400 epochs training recipe.
When applying RoPE to ViT, we remove APE from ViT by default. 
Thus, 2D RoPE is the only position embedding for RoPE ViT.
We denote ViT uses both RoPE and APE as RoPE+APE.

In Fig.~\ref{fig:exp_multi_vit}, we compare 2D RoPE variants with APE for ViT position embedding.
Both 2D RoPE, RoPE-Axial, and RoPE-Mixed implementations outperform APE for resolutions larger than 224, i.e., extrapolation cases.
As expected, the strong extrapolation performance of RoPE can be extended to image recognition tasks.
In comparison between RoPE-Axial and RoPE-Mixed, RoPE-Mixed performs better than RoPE-Axial in all input resolutions, meaning learnable frequencies for mixed axes are beneficial for classification. 

We measure the performance of RoPE-Mixed when it is used with APE.
The left side of Fig.~\ref{fig:exp_multi_RoPE_with} shows the performance of RoPE-Mixed with APE (RoPE-Mixed + APE) compared to RoPE-Mixed and APE. 
Note that we report accuracy improvement over APE for RoPE models to improve visualization.
When used with RoPE, APE is beneficial for interpolation (res $< 224$) but reduces improvement on extrapolation (res $> 224$).
RoPE+APE is almost double the improvement of RoPE-Mixed in interpolation, while the disadvantage in extrapolation is comparably small. 
Thus, RoPE+APE is a considerable choice for applying RoPE to ViT-based architectures on the target resolution of the tasks.

\vspace{-1em}
\subsubsection{Swin Transformer}

\begin{figure}[t]
    \centering
    \includegraphics[width=0.95\linewidth]{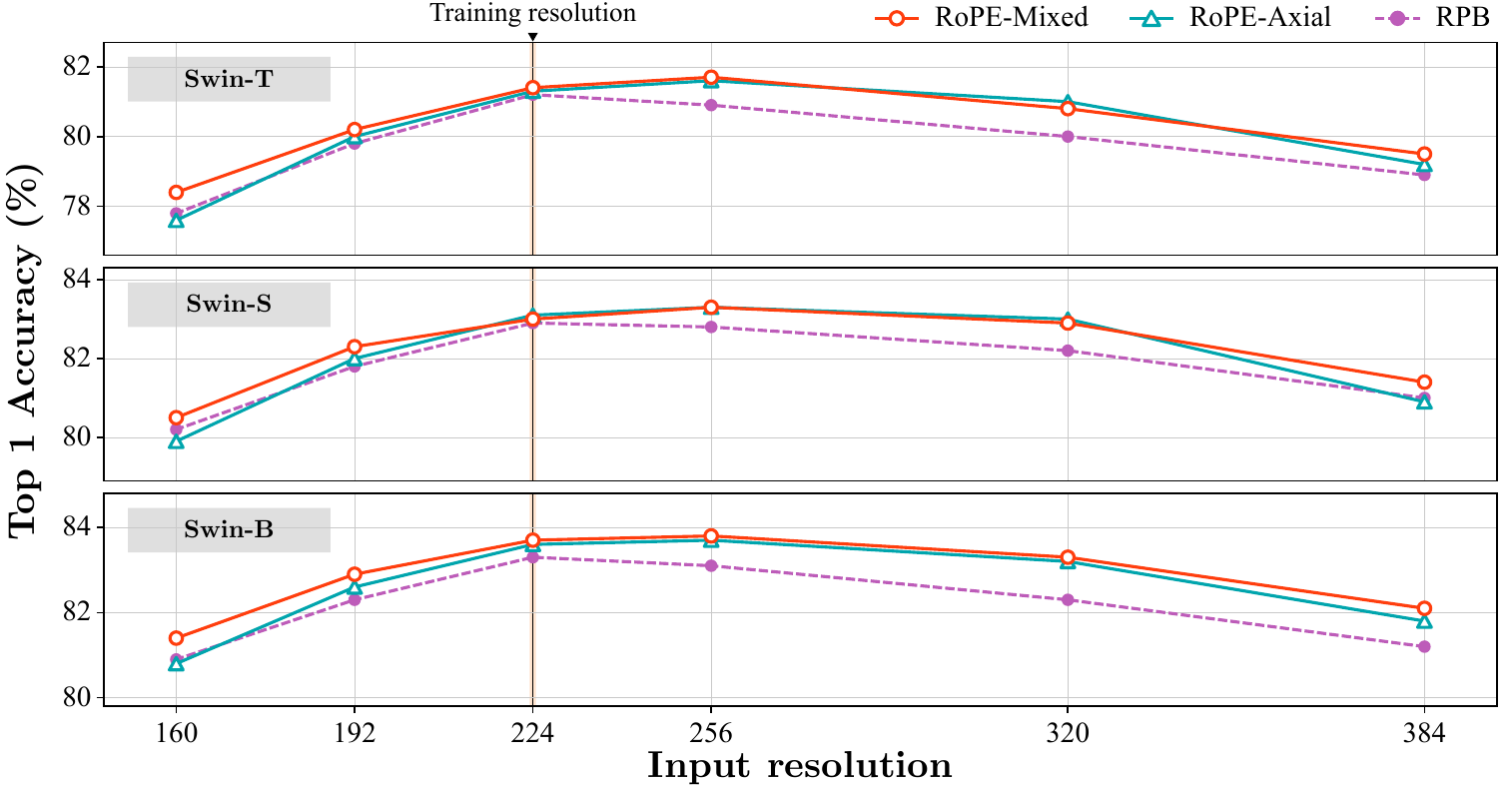}
    \vspace{-1em}
    \caption{\small \textbf{Multi-resolution performance of Swin Transformers.} We replace RPB in Swin Transformers with 2D RoPE variants: RoPE-Axial and RoPE-Mixed. Various Swin Transformers are trained with their 300 epochs training recipe~\cite{liu2021swin}. For multi-resolution inference, we change the window size of the window attention.}
    \label{fig:exp_multi_swin}
    \vspace{-1em}
\end{figure}

\begin{figure}[t]
    \centering
    \includegraphics[width=1.0\linewidth]{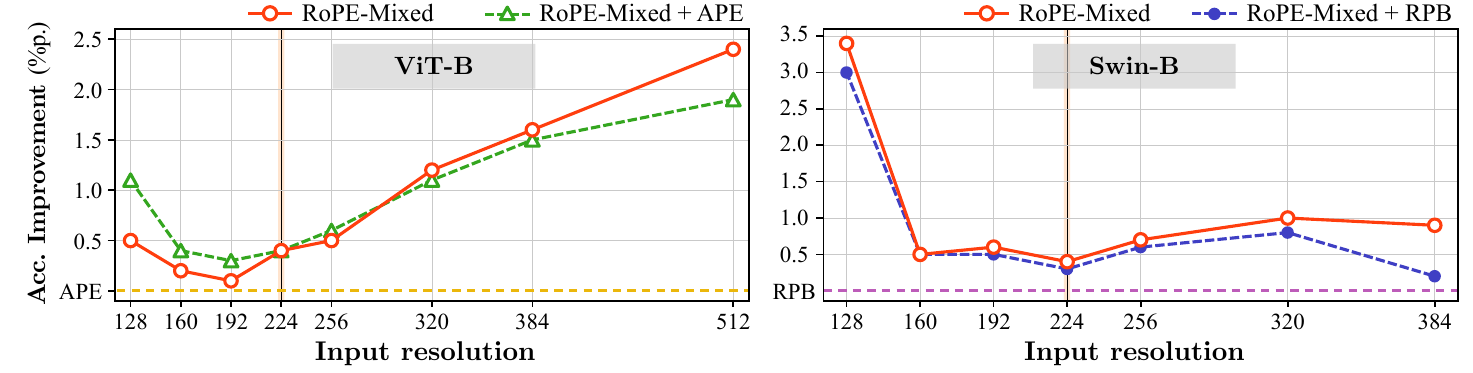}
    \vspace{-2em}
    \caption{\small \textbf{RoPE with conventional position embedding.} We report multi-resolution performance when RoPE is used with conventional position embeddings: APE and RPB. Performance improvement over baseline is reported to improve visualization.}
    \label{fig:exp_multi_RoPE_with}
    \vspace{-1em}
\end{figure}

2D RoPE variants are applied to Swin Transformers, a milestone work in hierarchical ViT with relative position embedding RPB.
The experiment in Swin Transformer investigates whether RoPE can replace RPB or work efficiently in a hierarchical ViT.
We train Swin-T, Swin-S, and Swin-B on ImageNet-1k with 300 epochs of Swin Transformer training recipe~\cite{liu2021swin}.
Similar to ViT, we replace RPB with 2D RoPE for comparison.
Thus, RoPE Swin (\ie Swin Transformer armed with RoPE) does not use RPB by default.
A Swin Transformer using both position embedding is dubbed RoPE+RPE.

Fig~\ref{fig:exp_multi_swin} shows the multi-resolution performance of various Swin Transformers with different position embeddings.
Two variants of 2D RoPE show remarkable performance improvements for extrapolation cases (res $> 224$).
Even in interpolation (res $< 224$), RoPE-Mixed outperforms RPB by a large margin.
It means that RoPE-Mixed is a more suitable option than RPB for Swin Transformers.
When comparing RoPE-Mixed with RoPE-Axial, RoPE-Mixed outperforms in most resolutions.
RoPE-Axial is especially weak in interpolation and significant extrapolation (res$=384$) cases.

We also measure performance when RoPE-Mixed is used together with RPB.
The right side of Fig.~\ref{fig:exp_multi_RoPE_with} shows the results.
Different from RoPE+APE in ViT, RoPE+RPB has no performance advantage compared to RoPE-Mixed in all resolutions.
This implies that RoPE-Mixed effectively replaces RPB as a relative position embedding.
Note that the gap between RoPE-Mixed and RoPE+RPB is significant when input resolution is far different from training resolution, demonstrating the advantage of RoPE-Mixed on resolution changes.

\subsection{Object detection}
\label{subsec:exp_det}

\begin{table}[t]
\centering
\caption{\textbf{MS-COCO object detection with DINO-ViTDet.} The table shows MS-COCO~\cite{lin2014mscoco} detection performance (box AP). DINO~\cite{zhang2022dino} is trained with DINO-ViTDet 12 epochs setting~\cite{ren2023detrex}. RoPE is applied to the backbone ViT, which is pre-trained on ImageNet-1k with DeiT-III 400epochs recipe.}
\vspace{-0.3cm}
\setlength\tabcolsep{8pt}
\begin{tabular}{@{}lccccc@{}}
\toprule
\multirow{2}{*}[-0.3em]{Backbone}  & \multirow{2}{*}[-0.3em]{APE}     & \multicolumn{4}{c}{RoPE}  \\ \cmidrule(l){3-6} 
     &        & Axial & Mixed & Axial+APE & Mixed+APE \\ \midrule
ViT-B & 49.4 & 50.8\greenpscript{+1.4}  & \textbf{51.2\greenpscript{+1.8}}   & 50.7\greenpscript{+1.3} & 51.1\greenpscript{+1.7}   \\
ViT-L & 51.1 & 52.2\greenpscript{+1.1}   & \textbf{52.9\greenpscript{+1.8}}   & 52.5\greenpscript{+1.4}     & 52.8\greenpscript{+1.7}             \\ \bottomrule
\end{tabular}
\label{table:det_vit}
\vspace{-0.5em}
\end{table}
\begin{table}[t]
\centering
\caption{\textbf{MS-COCO object detection with DINO-Swin.} MS-COCO~\cite{lin2014mscoco} detection performance (box AP) is reported for Swin Transformer with RoPE. DINO~\cite{zhang2022dino} is trained with DINO Swin 12 epochs setting~\cite{ren2023detrex}. Swin Transformers with RoPE or RPE are pre-trained on ImageNet-1k with Swin Transformer 300epochs recipe.}
\vspace{-0.3cm}
\setlength\tabcolsep{8pt}
\begin{tabular}{@{}lccccc@{}}
\toprule
\multirow{2}{*}[-0.3em]{Backbone}  & \multirow{2}{*}[-0.3em]{RPB}     & \multicolumn{4}{c}{RoPE}  \\ \cmidrule(l){3-6} 
     &        & Axial & Mixed & Axial+RPB & Mixed+RPB \\ \midrule
Swin-T & 51.3 & 51.6\greenpscript{+0.3}  & \textbf{51.8\greenpscript{+0.5}} & 51.7\greenpscript{+0.4} & 51.6\greenpscript{+0.3}   \\
Swin-S & 53.0 & 53.1\greenpscript{+0.1}  & 53.3\greenpscript{+0.3} & 53.5\greenpscript{+0.5}  & \textbf{53.6\greenpscript{+0.6}}   \\
Swin-B & 54.2 & 54.4\greenpscript{+0.2}   & 54.5\greenpscript{+0.3}  & \textbf{54.7\greenpscript{+0.5}}     & 54.5\greenpscript{+0.3}  \\ \bottomrule
\end{tabular}
\label{table:det_swin}
\vspace{-0.5em}
\end{table}

We verify 2D RoPE in object detection on MS-COCO~\cite{lin2014mscoco}. DINO~\cite{zhang2022dino} detector is trained using ViT and Swin as backbone network. We use ImageNet-1k weights from \S\ref{subsec:exp_multi_res} for pre-trained weights, and RoPE is only applied to the backbone. We use Detrex~\cite{ren2023detrex} codebase for detection training. DINO-ViTDet 12 epochs setting and DINO-Swin 12 epochs setting are used for DINO training.

Table~\ref{table:det_vit} shows the DINO-ViTDet results in bounding box AP. We report four variants of RoPEs: Axial, Mixed, Axial+APE, and Mixed+APE; all demonstrate remarkable performance improvements. DINO-ViTDet achieves AP improvement of more than +1.0pp by changing positional embedding to RoPE. Among RoPE variants, RoPE-Mixed shows the best improvement at +1.8pp. AP in ViT-B and ViT-L. DINO-ViTDet uses ViT backbone with window-block attention, but still, a few layers remain as global attention. We believe that RoPE is highly effective due to the extrapolation in global attention.

The performance of DINO-Swin is reported in Table~\ref{table:det_swin}. Like DINO-ViTDet, four variants are reported: Axial, Mixed, Axial+RPB, and Mixed+RPB. RoPE outperforms RPB for all variants. RoPE-Mixed performs better than RoPE-Axial. +RPB is beneficial for Axial but has limited effect on Mixed.
Performance improvement is smaller than DINO-ViTDet since DINO-Swin maintains a window size of the pre-trained backbone, i.e., DINO-Swin has no extrapolation. However, RoPE achieves meaningful gains and has room for improvement by increasing the Swin Transformer's window size for the detection backbone.

\vspace{-1em}
\subsection{Semantic segmentation}
\label{subsec:exp_seg}

\begin{table}[t]
\centering
\caption{\textbf{ADE20k semantic segmentation using the UperNet~\cite{xiao2018uppernet} head.} UperNet is trained with ViT backbone following ViT training recipe~\cite{peng2022beit2}. The table reports performance as mIoU metric. We report single-scale and multi-scale evaluation results.}
\vspace{-0.3cm}
\setlength\tabcolsep{8pt}
\begin{tabular}{@{}lcccccc@{}}
\toprule
 & \multirow{2}{*}[-0.3em]{\begin{tabular}{@{}c@{}}Multi-\\ scale\end{tabular}} & \multirow{2}{*}[-0.3em]{APE}     & \multicolumn{4}{c}{RoPE}  \\ \cmidrule(l){4-7} 
     &  &       & Axial & Mixed & Axial+APE & Mixed+APE \\ \midrule
\multirow{2}{*}[-0.1em]{ViT-B} & - & 47.7 & 49.0\greenpscript{+1.3}  & 49.6\greenpscript{+1.9}   & 49.5\greenpscript{+1.8} & \textbf{50.0\greenpscript{+2.3}}   \\
 & \textcolor{darkergreen}{\ding{52}} & 48.4 & 49.9\greenpscript{+1.5}  & 50.7\greenpscript{+2.3}  & 50.5\greenpscript{+2.1} & \textbf{50.9\greenpscript{+2.5}}   \\ \midrule
\multirow{2}{*}[-0.1em]{ViT-L} & - & 50.8 & 51.8\greenpscript{+1.0}   & 51.5\greenpscript{+0.7}  & 51.6\greenpscript{+0.8}      & \textbf{52.0\greenpscript{+1.2}}  \\ 
 & \textcolor{darkergreen}{\ding{52}} & 51.6 & \textbf{52.6\greenpscript{+1.0}}   & 52.3\greenpscript{+0.7}    & 52.4\greenpscript{+0.8}    & \textbf{52.6\greenpscript{+1.0}}  \\ \bottomrule
\end{tabular}
\label{table:seg_vit}
\vspace{-0.5em}
\end{table}
\begin{table}[t]
\centering
\caption{\textbf{ADE20k semantic segmentation with Swin-Mask2Former~\cite{cheng2021mask2former}.} Mask2Former model for semantic segmentation is trained using Swin Transformer. The table shows segmentation performance in mIoU metric.}
\vspace{-0.3cm}
\setlength\tabcolsep{8pt}
\begin{tabular}{@{}lccccc@{}}
\toprule
\multirow{2}{*}[-0.3em]{Backbone}  & \multirow{2}{*}[-0.3em]{RPB}     & \multicolumn{4}{c}{RoPE}  \\ \cmidrule(l){3-6} 
     &        & Axial & Mixed & Axial+RPB & Mixed+RPB \\ \midrule
Swin-S & 50.2 & 50.4\greenpscript{+0.2}  & 51.1\greenpscript{+0.9} & \textbf{51.2\greenpscript{+1.0}} & 50.9\greenpscript{+0.7}  \\
Swin-B & 51.5 & \textbf{52.0\greenpscript{+0.5}}   & \textbf{52.0\greenpscript{+0.5}} & 50.0\redpscript{-1.5} & 51.4\redpscript{-0.1}  \\ \bottomrule
\end{tabular}
\label{table:seg_swin}
\vspace{-.75em}
\end{table}

We train 2D RoPE ViT and Swin for semantic segmentation on ADE20k~\cite{zhou2017ade1,zhou2019ade2}.
For ViT, we use UperNet~\cite{xiao2018uppernet} with ViT training recipe~\cite{peng2022beit2}.
For Swin, Mask2Former~\cite{cheng2021mask2former} for segmentation is used with the Swin.
ImageNet-1k pre-trained weights from \S\ref{subsec:exp_multi_res} are used for pre-trained weights.
Also, RoPE is only applied to the backbone. The networks are trained for 160k iterations.

Table~\ref{table:seg_vit} shows ViT-UperNet performances. RoPE-based models achieve impressive performance improvement in all cases. It is noteworthy that Mixed+APE achieves +2.3 and +2.5 mIoU improvement with only position embedding changes. 
The improvement might originate from the extrapolation performance of RoPE since the ViT-UperNet setting uses $512 \times 512$ images for inputs.
Among the three variants of RoPE, Mixed+APE shows the best performance in all cases, which is different from detection results. As shown in Fig.~\ref{fig:exp_multi_RoPE_with}, Mixed+APE has an advantage at interpolation while degrading performance at extrapolation. 
These results suggest that the use of APE in a RoPE-based ViT should be adjusted based on the target task.
Swin-Mask2Former performances are shown in Table~\ref{table:seg_swin}. RoPE also improves the performance of Swin-based segmentation. RoPE-Mixed shows impressive performance, while +RPB is only beneficial in limited cases.

\subsection{Comparison with multi-resolution methods}

\begin{table}[t]
\centering
\caption{\textbf{Multi-resolution comparison with ResFormer~\cite{tian2023resformer}.} The table shows a comparison of RoPE-Mixed based ViTs with ResFormer-S trained for $224 \times 224 $ resolution. RoPE ViT outperforms ResFormer on extrapolation, resolution $> 224$, and shows comparable performance at small resolutions. }
    \vspace{-0.3cm}
\setlength\tabcolsep{4.5pt}
\begin{tabular}{@{}lcccccccccc@{}}
\toprule
Test resolution & 96 & 128 & 160 & 192 & 224 & 288 & 384 & 448 & 512 \\ 
\midrule
ResFormer-S & 57.8 & \textbf{71.4} & \textbf{77.0} & \textbf{79.6} & 80.8 & 81.4 & 80.7 & 79.3 & 77.7  \\
ViT-S  & 35.4 & 69.3 & 76.1 & 79.1 & 80.4 & 80.9 & 79.4 & 77.6 & 75.4  \\
+ RoPE-Mixed & 55.7 & 70.6 & 76.6 & \textbf{79.6} & \textbf{80.9} & 82.0 & \textbf{81.8} & \textbf{80.9} & \textbf{79.1}  \\
+ RoPE-M + APE & \textbf{58.5} & \textbf{71.4} & 76.7 & 79.5 & \textbf{80.9} & \textbf{82.3} & 81.7 & 80.5 & 78.5  \\
\bottomrule
\end{tabular}
\label{table:compare_reformer}
\vspace{-1.25em}
\end{table}

We compare 2D RoPE variants with recent ViT architecture designed for multi-resolution inference, namely ResFormer~\cite{tian2023resformer}. ResFormer uses depth-wise convolutions as the position embedding. It uses sinusoidal APE in Eq.~\ref{eq:sin_APE} and depth-wise convolution after the patch-embed layer as Global Position Embedding (GPE). Also, another depth-wise convolution is used similar to skip-connection for every self-attention layer to add position embed as Local Position Embed (LPE). Using GPE and LPE, ResFormer is proposed as an improved ViT for multi-resolution inference. ResFormer is trained with multi-resolution training utilizing self-distillation loss. Since self-distillation with multi-resolution training is not a common recipe in ViT, we use ResFormer-S trained with fixed resolution $224 \times 224$ and compare it with RoPE-Mixed ViT-S in \S\ref{subsec:exp_multi_res}. Table~\ref{table:compare_reformer} shows a multi-resolution comparison of RoPE-Mixed with ResFormer-S-224. RoPE-Mixed outperforms ResFormer with a meaningful margin for extrapolation ranges (res $> 224$), but RoPE-Mixed shows performance lower than ResFormer for significant interpolation ranges (res $\leq 160$). To achieve comparable interpolation, RoPE-Mixed needs additional APE. Overall, the results show that RoPE-Mixed+APE outperforms ResFormer-S in multi-resolution inference.


\vspace{-.5em}
\section{Conclusion}
\label{sec:conclusion}
\vspace{-.5em}
Rotary Position Embedding (RoPE) is a novel method for relative position embedding with a lot of potential. However, it has been underexplored in vision modeling. In this paper, we have conducted a comprehensive investigation of 2D RoPE for Vision Transformer (ViT) and proposed an improved 2D RoPE, RoPE-Mixed, utilizing mixed axis frequency with learnable parameters. Our experiments show that 2D RoPE is an effective solution for multi-resolution classification for both ViT and Swin Transformers, particularly for large resolutions. 2D RoPE shows improved performance with a significant margin in downstream tasks, such as object detection and semantic segmentation. It is noteworthy that our RoPE-Mixed outperforms conventional 2D RoPE in various tasks, further enhancing the contribution of this research. We believe that our study will be useful for vision researchers looking for state-of-the-art performance by suggesting 2D RoPE as a solution for them. 

%
%
\bibliographystyle{splncs04}
\bibliography{main}

\clearpage
\section*{Appendix}
\appendix
\numberwithin{equation}{section}
\numberwithin{figure}{section}
\numberwithin{table}{section}

\section{Experiments (cont'd)}

We demonstrated the performance of 2D RoPE with performance graphs through various input resolutions in the main paper. This appendix provides additional ablation studies and the entire performance numbers for the multi-resolution experiments. We measured the performance of ViTs and Swin Transformers with default position embedding (APE or RPB), 2D RoPE variants (RoPE-Axial and RoPE-Mixed), and 2D RoPE variants with default position embedding (RoPE-Axial and RoPE-Mixed + APE or RPB). Note that figures in the paper do not include small resolutions such as $96 \times 96$ to improve the visualization.

\subsection{Impacts of learnable frequencies}
Studies on applying RoPE to ViT~\cite{fang2023eva,lu2023unified,lu2024fit} have not considered the learnable frequencies. 
However, a comparison with RoPE-Axial + learnable frequencies can be an interesting ablation study by revealing the contribution of learnable frequencies on RoPE-Mixed.
Table~\ref{rtable:axial_learn_s} and Table~\ref{rtable:axial_learn_b} show learnable RoPE-Axial performance compared to fixed RoPE-Axial and RoPE-Mixed. The learnable frequencies improve RoPE-Axial on high-resolution (384, 512) but are ineffective on other resolutions. These results imply that the effects of RoPE-Mixed originate from frequency mixing rather than frequency learning, as we claimed in the paper.
\begin{table}[h]
    \vspace{-1em}
    \centering
\setlength\tabcolsep{5pt}
    \caption{RoPE-Axial with learnable frequencies for ViT-S.}
    \begin{tabular}{lccccccc}
        \toprule
Resolution       & 144     & 192     & 224     & 256     & 320     & 384     & 512     \\
\midrule
Axial       & 73.6    & 79.2    & 80.9    & 81.7    & 81.5    & 80.0    & 76.1    \\
\midrule
\multirow{2}{*}{Axial+learn} & 73.5  & 79.1    & 80.7    & 81.5    & 81.8    & 81.3    & 77.8    \\
                 & \redp{-0.1}  & \redp{-0.1}    & \redp{-0.2}    & \redp{-0.2}    & \greenp{+0.3}     & \greenp{+1.3}     & \greenp{+1.7}     \\
\midrule
\multirow{2}{*}{Mixed}    & 74.2    & 79.6    & 80.9    & 81.8    & 82.2    & 81.8    & 79.1    \\
              & \greenp{+0.6}     & \greenp{+0.4}     & (0.0)     & \greenp{+0.1}     & \greenp{+0.7}     & \greenp{+1.8}     & \greenp{+3.0}     \\
        \bottomrule
    \end{tabular}
    \vspace{-1em}
    \label{rtable:axial_learn_s}
    \vspace{-2em}
\end{table}
\begin{table}[h]
    \centering
\setlength\tabcolsep{5pt}
    \caption{RoPE-Axial with learnable frequencies for ViT-B.}
    \begin{tabular}{lccccccc}
        \toprule
Resolution       & 144     & 192     & 224     & 256     & 320     & 384     & 512     \\
\midrule
Axial       & 78.9    & 82.8    & 83.6    & 84.2    & 84.3    & 83.9    & 82.0    \\
\midrule
\multirow{2}{*}{Axial+learn}   & 78.9    & 82.7    & 83.6    & 84.2    & 84.4    & 83.9    & 82.3    \\
          & (0.0)     & \redp{-0.1}    & (0.0)     & (0.0)     & \greenp{+0.1}     & 0.0     & \greenp{+0.3}     \\
\midrule
\multirow{2}{*}{Mixed}        & 79.4    & 82.8    & 83.8    & 84.3    & 84.7    & 84.4    & 82.9    \\
          & \greenp{+0.5}     & (0.0)     & \greenp{+0.2}     & \greenp{+0.1}     & \greenp{+0.4}     & \greenp{+0.5}     & \greenp{+0.9}     \\
        \bottomrule
    \end{tabular}
    \vspace{-1em}
    \label{rtable:axial_learn_b}
    \vspace{-1em}
\end{table}

\subsection{Multi-resolution classification -- ViT}
Table~\ref{atab:vit_s}, \ref{atab:vit_b}, and \ref{atab:vit_l} report the total numbers of multi-resolution classification, which is illustrated in Figure \ref{fig:exp_multi_vit}
. 2D RoPE variants outperform APE in the smallest resolution $96 \times 96$ with significant gap. 

\begin{table}[H]
\centering 
\small
\setlength\tabcolsep{5pt}
\caption{\textbf{Multi-resolution performance of ViT-S.} Table reports the performance of 2D RoPE variants corresponding to the first graph in Figure \ref{fig:exp_multi_vit}
.}
\resizebox{1.0\linewidth}{!}{
\begin{tabular}{lcccccccc}
\toprule
 & \multicolumn{8}{c}{Test resolution} \\
\cmidrule{2-9}
Position embeds & 96$\times$96 & 144$\times$144 & 192$\times$192 & 224$\times$224 & 256$\times$256 & 320$\times$320 & 384$\times$384 & 512$\times$512 \\
\midrule
APE & 35.4 & 73.6 & 79.1 & 80.4 & 80.9 & 80.6 & 79.4 & 75.4 \\ \midrule
RoPE-Axial & 55.9 & 73.6 & 79.2 & 80.9 & 81.7 & 81.5 & 80.0 & 76.1 \\
RoPE-Mixed & 55.7 & 74.2 & 79.6 & 80.9 & 81.8 & 82.2 & 81.8 & 79.1 \\ \midrule
APE + RoPE-Axial & 58.4 & 74.2 & 79.2 & 80.7 & 81.6 & 81.9 & 81.2 & 75.3 \\
APE + RoPE-Mixed & 58.5 & 74.4 & 79.5 & 80.9 & 81.8 & 82.1 & 81.7 & 78.5 \\
\bottomrule
\end{tabular}
}
\label{atab:vit_s}
\end{table}

\begin{table}[H]
\centering
\small
\setlength\tabcolsep{5pt}
\caption{\textbf{Multi-resolution performance of ViT-B.} Table reports the performance of 2D RoPE variants corresponding to the second graph in Figure \ref{fig:exp_multi_vit}
.}
\resizebox{1.0\linewidth}{!}{
\begin{tabular}{lcccccccc}
\toprule
& \multicolumn{8}{c}{Test resolution} \\ \cmidrule{2-9}
Position embeds & 96$\times$96 & 144$\times$144 & 192$\times$192 & 224$\times$224 & 256$\times$256 & 320$\times$320 & 384$\times$384 & 512$\times$512 \\ \midrule
APE & 57.6 & 79.1 & 82.7 & 83.4 & 83.8 & 83.5 & 82.8 & 80.5 \\ \midrule
RoPE-Axial & 66.9 & 78.9 & 82.8 & 83.6 & 84.2 & 84.3 & 83.9 & 82.0 \\
RoPE-Mixed & 68.1 & 79.4 & 82.8 & 83.8 & 84.3 & 84.7 & 84.4 & 82.9 \\ \midrule
APE + RoPE-Axial & 68.9 & 79.3 & 82.8 & 83.7 & 84.2 & 84.4 & 83.8 & 81.4 \\
APE + RoPE-Mixed & 70.2 & 79.7 & 83.0 & 83.8 & 84.4 & 84.6 & 84.3 & 82.4 \\
\bottomrule
\end{tabular}
}
\label{atab:vit_b}

\end{table}

\begin{table}[H]
\centering
\small
\setlength\tabcolsep{5pt}
\caption{\textbf{Multi-resolution performance of ViT-L.} Table reports the performance of 2D RoPE variants corresponding to the third graph in Figure \ref{fig:exp_multi_vit}
.}
\resizebox{1.0\linewidth}{!}{
\begin{tabular}{lcccccccc}
\toprule
& \multicolumn{8}{c}{Test resolution} \\ \cmidrule{2-9}
Position embeds & 96$\times$96 & 144$\times$144 & 192$\times$192 & 224$\times$224 & 256$\times$256 & 320$\times$320 & 384$\times$384 & 512$\times$512 \\ \midrule
APE & 61.5 & 80.8 & 83.8 & 84.6 & 84.9 & 84.7 & 84.2 & 82.2 \\ \midrule
RoPE-Axial & 71.0 & 80.9 & 83.9 & 84.7 & 85.1 & 85.3 & 85.1 & 84.0 \\
RoPE-Mixed & 71.7 & 81.1 & 83.9 & 84.8 & 85.4 & 85.7 & 85.6 & 84.7 \\ \midrule
APE + RoPE-Axial & 72.4 & 81.1 & 84.0 & 84.7 & 85.2 & 85.3 & 85.1 & 83.8 \\
APE + RoPE-Mixed & 73.2 & 81.3 & 84.0 & 84.9 & 85.3 & 85.6 & 85.5 & 84.4 \\
\bottomrule
\end{tabular}
}
\label{atab:vit_l}

\end{table}

\subsection{Multi-resolution classification -- Swin Transformer}

Table~\ref{atab:swin_t}, \ref{atab:swin_s}, and \ref{atab:swin_b} show the total numbers of multi-resolution classification of Swin Transformer with 2D RoPE variants corresponding to Figure \ref{fig:exp_multi_swin} 
of paper. Similar to ViT cases, 2D RoPE variants significantly outperform RPB in small resolutions: $96\times 96$ and $128 \times 128$.

\begin{table}[H]
\centering
\small
\setlength\tabcolsep{5pt}
\caption{\textbf{Multi-resolution performance of Swin-T.} Table reports Swin-T performance with 2D RoPE variants corresponding to the first graph in Figure \ref{fig:exp_multi_swin}
.}
\resizebox{1.0\linewidth}{!}{
\begin{tabular}{lcccccccc}
\toprule
& \multicolumn{8}{c}{Test resolution} \\ \cmidrule{2-9}
Position embeds & 96$\times$96 & 128$\times$128 & 160$\times$160 & 192$\times$192 & 224$\times$224 & 256$\times$256 & 320$\times$320 & 384$\times$384 \\ \midrule
RPB & 39.6 & 67.4 & 77.8 & 79.8 & 81.2 & 80.9 & 80.0 & 78.9  \\ \midrule
RoPE-Axial & 47.6 & 69.5 & 77.6 & 80.0 & 81.3 & 81.6 & 81.0 & 79.2  \\
RoPE-Mixed & 53.6 & 71.9 & 78.4 & 80.2 & 81.4 & 81.7 & 80.8 & 79.5   \\ \midrule
RPB + RoPE-Axial & 43.8 & 67.9 & 77.8 & 80.2 & 81.5 & 81.6 & 80.9 & 79.1   \\
RPB + RoPE-Mixed & 50.9 & 69.4 & 78.1 & 80.3 & 81.5 & 81.8 & 80.7 & 78.5  \\ \bottomrule
\end{tabular}
}
\label{atab:swin_t}
\end{table}

\begin{table}[H]
\centering
\small
\setlength\tabcolsep{5pt}
\caption{\textbf{Multi-resolution performance of Swin-S.} Table reports Swin-S performance with 2D RoPE variants corresponding to the second graph in Figure \ref{fig:exp_multi_swin}
.}
\resizebox{1.0\linewidth}{!}{
\begin{tabular}{lcccccccc}
\toprule
& \multicolumn{8}{c}{Test resolution} \\ \cmidrule{2-9}
Position embeds & 96$\times$96 & 128$\times$128 & 160$\times$160 & 192$\times$192 & 224$\times$224 & 256$\times$256 & 320$\times$320 & 384$\times$384 \\ \midrule
RPB & 47.0 & 72.7 & 80.2 & 81.8 & 82.9 & 82.8 & 82.2 & 81.0 \\ \midrule
RoPE-Axial & 44.0 & 72.0 & 79.9 & 82.0 & 83.1 & 83.3 & 83.0 & 80.9 \\
RoPE-Mixed & 55.7 & 75.5 & 80.5 & 82.3 & 83.0 & 83.3 & 82.9 & 81.4 \\ \midrule
RPB + RoPE-Axial & 55.4 & 74.7 & 80.8 & 82.4 & 83.2 & 83.3 & 82.8 & 81.3 \\
RPB + RoPE-Mixed & 57.4 & 75.2 & 80.8 & 82.5 & 83.3 & 83.4 & 82.8 & 81.1 \\ \bottomrule
\end{tabular}
}
\label{atab:swin_s}

\end{table}

\begin{table}[H]

\centering
\small
\setlength\tabcolsep{5pt}
\caption{\textbf{Multi-resolution performance of Swin-B.} Table reports Swin-B performance with 2D RoPE variants corresponding to the third graph in Figure \ref{fig:exp_multi_swin}
.}
\resizebox{1.0\linewidth}{!}{
\begin{tabular}{lcccccccc}
\toprule
& \multicolumn{8}{c}{Test resolution} \\ \cmidrule{2-9}
Position embeds & 96$\times$96 & 128$\times$128 & 160$\times$160 & 192$\times$192 & 224$\times$224 & 256$\times$256 & 320$\times$320 & 384$\times$384 \\ \midrule
RPB & 48.8 & 73.3 & 80.9 & 82.3 & 83.3 & 83.1 & 82.3 & 81.2 \\ \midrule
RoPE-Axial & 52.7 & 74.3 & 80.8 & 82.6 & 83.6 & 83.7 & 83.2 & 81.8 \\
RoPE-Mixed & 61.5 & 76.7 & 81.4 & 82.9 & 83.7 & 83.8 & 83.3 & 82.1 \\ \midrule
RPB + RoPE-Axial & 55.3 & 74.4 & 81.3 & 82.8 & 83.6 & 83.8 & 83.1 & 81.5 \\
RPB + RoPE-Mixed & 62.2 & 76.3 & 81.4 & 82.8 & 83.6 & 83.7 & 83.1 & 81.4 \\ \bottomrule
\end{tabular}
}
\label{atab:swin_b}

\end{table}

\subsection{2D RoPE with APE or RPB}
In Figure \ref{fig:exp_multi_RoPE_with} 
of the paper, we report the performance improvement of RoPE-Mixed compared to base position embeddings: APE or RPB. We provide numbers for Figure \ref{fig:exp_multi_RoPE_with} 
in Table~\ref{atab:vit_improv} and \ref{atab:swin_improv}. Note that each number means performance improvement (\%p.) compared to base position embeddings (APE or RPB). 

\begin{table}[h]

\centering
\small
\setlength\tabcolsep{5pt}
\caption{\textbf{Performance improvement compared to APE in ViT-B.} Table shows the improvement over APE, which is shown in the left graph of Figure \ref{fig:exp_multi_RoPE_with}
.}
\resizebox{1.0\linewidth}{!}{
\begin{tabular}{lccccccccc}
\toprule
& \multicolumn{9}{c}{Test resolution} \\ \cmidrule{2-10}
Position embeds & 112$\times$112 & 128$\times$128 & 160$\times$160 & 192$\times$192 & 224$\times$224 & 256$\times$256 & 320$\times$320 & 384$\times$384 & 512$\times$512 \\ \midrule
APE & 0.0 & 0.0 & 0.0 & 0.0 & 0.0 & 0.0 & 0.0 & 0.0 & 0.0 \\ \midrule
RoPE-Mixed & 1.1 & 0.5 & 0.2 & 0.1 & 0.4 & 0.5 & 1.2 & 1.6 & 2.4 \\ 
RoPE-Mixed + APE & 2.2 & 1.1 & 0.4 & 0.3 & 0.4 & 0.6 & 1.1 & 1.5 & 1.9 \\ \bottomrule
\end{tabular}
}
\label{atab:vit_improv}
\end{table}

\begin{table}[h]
\centering
\small
\setlength\tabcolsep{5pt}
\caption{\textbf{Performance improvement compared to RPB in Swin-B.} Table shows the improvement over RPB, which is shown in the right graph of Figure \ref{fig:exp_multi_RoPE_with}
.}
\resizebox{1.0\linewidth}{!}{
\begin{tabular}{lccccccc}
\toprule
& \multicolumn{7}{c}{Test resolution} \\ \cmidrule{2-8}
Position embeds & 128$\times$128 & 160$\times$160 & 192$\times$192 & 224$\times$224 & 256$\times$256 & 320$\times$320 & 384$\times$384 \\ \midrule
RPB & 0 & 0 & 0 & 0 & 0 & 0 & 0 \\ \midrule
RoPE-Mixed & 3.4 & 0.5 & 0.6 & 0.4 & 0.7 & 1.0 & 0.9 \\
RoPE-Mixed + RPB & 3.0 & 0.5 & 0.5 & 0.3 & 0.6 & 0.8 & 0.2 \\
\bottomrule
\end{tabular}
}
\label{atab:swin_improv}
\end{table}
\end{document}